%% file: main_2026.tex
\documentclass{article}

 \usepackage[main, final]{neurips_2026}

\usepackage[utf8]{inputenc} 
\usepackage[T1]{fontenc}    
\usepackage{hyperref}       
\usepackage{url}            
\usepackage{booktabs}       
\usepackage{amsfonts}       
\usepackage{nicefrac}       
\usepackage{microtype}      
\usepackage{xcolor}         

\usepackage{wrapfig}
\usepackage{graphicx} 
\usepackage{caption}
\usepackage{subcaption}
\usepackage{amsmath}
\usepackage{multirow}
\usepackage{makecell}
\usepackage{soul}
\definecolor{darkgreen}{RGB}{34, 139, 34} 
\usepackage{threeparttable}

\title{When Machines Speak: A Unified Generative Framework for Integrating Machine-Native Symbols into Pretrained Large Language Models}

\author{%
  Su Yan \\
  Google Inc.\\
  \texttt{sueyan@google.com} \\
   \And
   Rakesh Iyer \\
  Google Inc. \\
  \texttt{rni@google.com} \\
}

\begin{document}

\maketitle

\begin{abstract}

\input{abstract}

\end{abstract}

\section{Introduction}

\input{introduction}
\section{Background and related work}
\input{related_work}

\section{UniLang framework}
\input{body}

\section{Experiment}
\label{experiment}
\input{experiment}

\section{Conclusion and future work}
\label{conclusion}
\input{conclusion}

\newpage
\appendix
\setcounter{section}{0}
\renewcommand{\thesection}{\Alph{section}}
\label{appendix}
\input{appendix}

\newpage
\bibliographystyle{plain}        
\bibliography{llm_internalization}  



\end{document}

%% file: abstract.tex
Many real-world AI systems represent entities, behaviors, and structured information using discrete machine-native symbols rather than natural language. While these representations are compact and preserve task-relevant structure, they lie outside the linguistic token space of pretrained large language models (LLMs), creating a fundamental divide between language modeling and structured prediction. We introduce \textit{UniLang}, a unified generative framework that bridges this divide by extending pretrained LLMs to treat machine-native symbols as first-class generative units alongside natural-language tokens. UniLang expands the LLM's vocabulary and embedding space with grounded machine-native representations, enabling textual and symbolic tokens to be jointly modeled and generated under a single autoregressive objective. This unified interface allows pretrained LLMs to directly operate on machine-native representations without requiring them to be verbalized as natural language or relying on task-specific architectures. We evaluate UniLang on two structurally distinct tasks, \textit{sequential recommendation} and \textit{legal precedent prediction}, spanning different domains and types of structured prediction. Across both tasks, UniLang consistently outperforms strong baselines, demonstrating a path toward extending pretrained LLMs beyond language and using them as a common generative modeling backbone for heterogeneous machine-native representations.

%% file: introduction.tex
Large language models (LLMs) have emerged as a general modeling framework across a wide range of applications. Their success stems from a unified autoregressive formulation in which diverse natural-language tasks are expressed through a common vocabulary and generation objective. However, this unified interface is fundamentally linguistic: pretrained LLMs operate over natural-language tokens.

Many modern AI systems, in contrast, represent information using \emph{machine-native symbolic representations} rather than natural language. These representations increasingly take the form of discrete codes produced by vector quantization or related techniques, representing complex entities such as compressed audio signals~\cite{zeghidour2021soundstream}, semantic identifiers in recommendation systems~\cite{hou2023_vq_rec, Tiger}, and structured relational information in graph learning~\cite{luo2025node}. Such representations preserve explicit discrete structure and offer computational efficiency that natural language cannot easily replicate. However, they remain outside the linguistic token space of pretrained LLMs, making them difficult for LLMs to natively model and generate.

This mismatch creates a methodological divide. Existing approaches largely follow one of two paradigms. The first treats \emph{natural language as a universal interface}, converting structured information into textual descriptions so that pretrained LLMs can process it~\cite{P5, Jiang2023HealthSystemScale, neurips2025_think_before_recommendation, Luo2024TextguidedSmallMolecule, Zhang2024CitaLaw, neurips2024_graphtext}. While effective, verbalization may obscure native structure, introduce representational ambiguity, and sacrifice machine-level precision. The second operates directly on machine-native representations using task-specific models~\cite{SASRec, Tiger, Linger, zhou2020_s3rec}. Although these approaches preserve structural fidelity, they do not directly leverage the linguistic and world knowledge encoded in a pretrained LLM within the same generative space.

At a deeper level, the limitation is representational. Pretrained LLMs lack a unified interface that allows machine-native symbols to function as \emph{first-class generative units} alongside natural-language tokens. Without such an interface, structured symbolic prediction and language modeling remain separate paradigms.

This observation motivates a simple question:

\begin{quote}
\emph{Can pretrained LLMs be extended to directly model and generate machine-native symbols, rather than forcing symbolic information to become natural language or abandoning pretrained language models altogether?}
\end{quote}

We answer this question with \textbf{UniLang}\footnote{Code is available at \url{https://github.com/Stella-S-Yan/llm-internalization}}, a unified framework that extends pretrained LLMs to jointly model natural-language tokens and machine-native symbols within a single autoregressive vocabulary and generation objective. Rather than treating machine-native representations as external objects or auxiliary embeddings, UniLang explicitly grounds them into the pretrained LLM representation space, allowing machine-native symbols and natural-language tokens to function as first-class generative units within the same autoregressive framework.

This unified representational interface enables structurally different prediction problems to be expressed under the same modeling framework without task-specific architectures. To demonstrate its generality, we evaluate UniLang on two structurally distinct tasks: \emph{sequential recommendation} and \emph{legal precedent prediction}. These tasks were intentionally selected because they represent heterogeneous structured prediction problems rather than multiple datasets from the same application domain. Across both tasks, UniLang consistently outperforms strong baselines while using the same modeling framework.

We make the following contributions:

\begin{itemize}
    \item \textbf{A unified representational interface.}
    We formulate structured prediction as typed autoregressive generation over heterogeneous token types, allowing natural-language tokens and machine-native symbols to function as first-class generative units within the same pretrained LLM.

    \item \textbf{Grounded integration of machine-native symbols.}
    We introduce a vocabulary-extension and contrastive grounding procedure that maps machine-native symbols into the pretrained LLM's representation space, enabling them to be processed and generated alongside natural-language tokens while retaining their discrete structure.

    \item \textbf{Generality across heterogeneous structured prediction tasks.}
    We demonstrate that the same UniLang framework applies to structurally distinct tasks, including sequential recommendation and legal precedent prediction, without task-specific architectural modifications, and consistently outperforms strong baselines.
\end{itemize}

%% file: related_work.tex
\definecolor{contextyellow}{RGB}{255, 243, 176}
\definecolor{quotepink}{RGB}{255, 200, 200}
\definecolor{citationblue}{RGB}{190, 220, 255}
\definecolor{contextlightyellow}{RGB}{255, 250, 230}
\definecolor{citationlightblue}{RGB}{230, 240, 255}

\newcommand{\hlc}[2]{\sethlcolor{#1}\hl{#2}}

\subsection{Machine-native representation.}
Machine-native representations convert natural-language content into compact, structured sequences of discrete symbols optimized for computation. The central idea is to encode rich semantic information in a concise symbolic form that facilitates efficient storage, retrieval, and reasoning.

A well-known example is the use of ICD codes in the medical domain \cite{Hirsch2016ICD10History}, where detailed diagnoses are represented as standardized symbolic identifiers. These codes capture complex, structured information that would otherwise require lengthy textual descriptions. Nowadays, machine-native representations follow the same principle but often are learned automatically from data rather than manually defined.

\begin{wrapfigure}{r}{0.5\textwidth} 
  \vspace{-10pt} 
  \centering
  \includegraphics[width=0.5\textwidth]{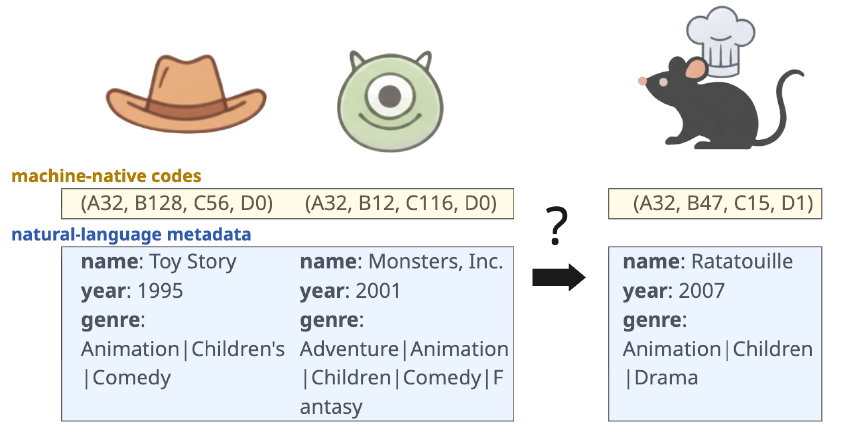}
  \captionsetup{font=small}
  \caption{Example of the sequential prediction task. Each item is collectively represented by \hlc{contextlightyellow}{machine-native codes} and various metadata in \hlc{citationlightblue}{natural language}.}
  \label{fig:sec_rec_example}
  \vspace{-10pt} 
\end{wrapfigure}

A variety of approaches have been proposed to generate such representations. Early methods rely on clustering-based discretization, such as hierarchical \textit{k}-means over embedding spaces \cite{tay2022transformermemorydifferentiablesearch}. Hashing-based techniques map high-dimensional continuous vectors into compact binary codes \cite{luo2025learninghashrecommendationsurvey}. Quantization-based approaches, including subspace quantization via \textit{k}-means, learn discrete codebooks for vector compression \cite{hou2023_vq_rec, jegou2011_product_quantization}. More recent work employs hierarchical residual quantization models, such as Residual Quantized Variational Autoencoders (RQ-VAE), to produce multi-level discrete codes that preserve semantic structure \cite{luo2025node, Tiger, Linger}. Figure \ref{fig:sec_rec_example} illustrates an example where each item (a movie) is associated with a discrete, hierarchical machine-native code generated by RQ-VAE. Details of the code construction procedure are provided in Section \ref{sec:constructing_machine_native_representation}.

\subsection{Sequential recommendation.} 
Sequential recommendation aims to predict the next item a user will interact with given their historical interaction sequence. A wide range of neural architectures have been explored for this task, including convolutional, recurrent, graph-based, and Transformer-based models \cite{moreira2021_transformers4rec, hidasi2015sessionbased, kang2018_sasrec,  li2017_neural_attentive_session, sun2019_bert4rec, tang2018_caser, zhou2020_s3rec}. 

Recent work has explored reformulating recommendation as a language modeling problem by converting user, item, and interaction information into unified natural-language sequences and training encoder-decoder models \cite{P5}. While this verbalization strategy enables the use of LLMs, flattening structured interactions into text can obscure type information and introduce representational ambiguity. 

Other approaches encode items as discrete symbols derived from vector quantization of dense embeddings~\cite{hou2023_vq_rec}. Extending this idea, \cite{Tiger} generates hierarchical discrete representations using Residual Quantized VAEs (RQ-VAE), referring to the resulting identifiers as Semantic IDs, and integrates them within generative retrieval models. Later work notes that purely discrete codes may lose fine-grained semantic information, and proposes augmenting them with continuous embeddings to recover these signals \cite{Linger}. While this hybrid approach mitigates information loss, it still treats discrete symbols indirectly within learned embedding spaces rather than as first-class generative units in a unified language modeling framework.

In contrast, UniLang models machine-native symbols and natural-language tokens together as generative units, enabling joint autoregressive prediction of structured and textual content.

\begin{wrapfigure}{r}{0.5\textwidth} 
  \vspace{-10pt} 
  \centering
  \includegraphics[width=0.5\textwidth]{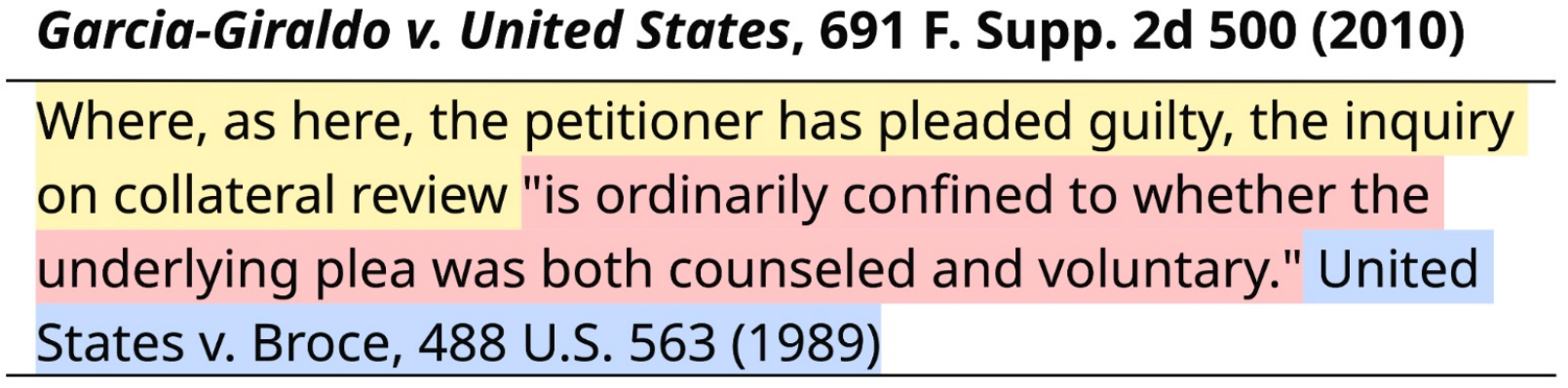}
  \captionsetup{font=small}
  \caption{Example of the legal precedent prediction task. Given \hlc{contextyellow}{context} from the \textbf{citing opinion} (\textit{Garcia-Giraldo v. United States}), predict the  \hlc{quotepink}{quotation} sentence(s) or paragraph(s) from the \hlc{citationblue}{cited opinion} (\textit{United States v. Broce}), which is unknown at inference time.}
  \label{fig:legal_example}
  \vspace{-10pt} 
\end{wrapfigure}

\subsection{Legal precedent prediction. }
Given the context of a legal argument, legal precedent prediction seeks to identify the specific sentence(s) or paragraph(s) from precedential court decisions that supports the claim \cite{mahari2021autolawaugmentedlegalreasoning}. This task is substantially more challenging than case-level citation prediction or document retrieval, where the objective is to identify a relevant case as a whole \cite{Dadgostari2021ModelingLawSearch, Huang2021ContextAwareLegalCitation}. Legal cases often span tens or hundreds of pages, requiring passage-level methods to perform fine-grained semantic alignment and more involved legal reasoning. Figure 2 illustrates an example citation context and its corresponding target passage from the legal dataset--LePaRD \cite{mahari2024_lepard}.

\subsection{Contrastive alignment and token grounding.}
To integrate non-linguistic data into language models, the primary challenge is bridging the semantic gap between human vocabulary and specialized machine symbols. Contrastive alignment has been widely used to ground heterogeneous representations into shared embedding spaces, most prominently in multimodal models such as CLIP \cite{CLIP}. Related techniques have also been explored for grounding new tokens or symbols into pretrained models \cite{gal2022imageworthwordpersonalizing, DBLP:journals/corr/abs-2010-06775}. 
Our work differs fundamentally in scope and purpose: rather than maintaining separate representational pathways or learning auxiliary embeddings, we use contrastive grounding as a mechanism to elevate machine-native discrete representations to first-class language tokens. This allows them to be generated, composed, and jointly modeled over within a unified sequence modeling framework.





%% file: body.tex
The core philosophy of UniLang is that natural language and machine-native representations are complementary representations of the same underlying entities. Natural language provides rich, human-interpretable context, while machine-native representations encode compact semantic structures optimized for computation. 

Unlike existing approaches that force a choice between these modalities, UniLang treats them as heterogeneous fields within a unified representation and generation space. By enabling joint computation over both types of information, UniLang combines the pretrained linguistic knowledge of LLMs with the compact semantic structure encoded by machine-native symbols.

\subsection{Constructing machine-native representations}
\label{sec:constructing_machine_native_representation}

UniLang is agnostic to the specific mechanism used to generate machine-native representations, requiring only that they consist of discrete, identifiable tokens that can be incorporated into the model vocabulary. In this work, following prior literature on Semantic IDs and residual vector quantization~\cite{Tiger, Linger}, we adopt an RQ-VAE–based discretization pipeline as a concrete instantiation.

Many items of interest admit partial textual descriptions. For example, a movie can be characterized by its title, release year, and genres, while a legal case is naturally described by its textual context, potentially augmented with structured metadata such as court name or case date. A pretrained text encoder maps these descriptions into high-dimensional embeddings, which an RQ-VAE discretizes into a sequence of codes across $l$ quantization levels. Denoting the codebook size at level $i$ by $c_i$, the resulting representation is $\mathbf{q} = (q^{(1)}, \dots, q^{(l)}), \quad q^{(i)} \in \{0, \dots, c_i - 1\}.$ Following common practice, we set $l=3$ and $c_i=256$ for all levels. To mitigate potential code collisions, we append an additional disambiguation level $q^{(l+1)}$, yielding $(q^{(1)}, \dots, q^{(l)}, q^{(l+1)})$. For example, $(11, 43, 204, 0)$ and $(11, 43, 204, 1)$ denote two distinct items sharing the same base quantized code. To integrate these discrete codes into the LLM vocabulary, we prepend level-specific prefixes to ensure token-level distinguishability, forming \textit{Semantic IDs} (SIDs) such as (A11, B43, C204, D0). The resulting machine-native vocabulary contains $4 \times 256 = 1{,}024$ tokens, which we call \textit{machine tokens}, capable of representing up to $256^4 \approx 4.3$ billion distinct items. Figure~\ref{fig:sec_rec_example} illustrates how items are jointly represented using SIDs and natural-language tokens.

\subsection{Vocabulary expansion and semantic grounding}
\label{sec:contrastive_loss}
To facilitate seamless modeling of non-linguistic symbols, UniLang establishes a shared vocabulary across modalities. Although pretrained LLMs are optimized for natural language, we extend their latent space with a dedicated set of machine tokens.

Machine tokens are grounded by aligning their embeddings with the textual description embeddings of corresponding items using contrastive learning. Given a pretrained LLM with natural-language vocabulary $\mathcal{V}_{\mathrm{NL}}$, we extend its vocabulary with a fixed set of machine tokens $\mathcal{V}_{\mathrm{ML}}$ (1,024 tokens in our instantiation), each assigned a learnable embedding initialized from a zero-mean normal distribution with standard deviation equal to the model's initializer range.

Let $\mathbf{m}_i = (m_i^1, m_i^2, \dots, m_i^L) \in \mathcal{V}_\mathrm{ML}^L$ denote the machine token sequence representing item $i$, where $L$ is fixed (e.g. $L=4$ in our setting). Let $\mathbf{t}_i = (w_i^1, w_i^2, \dots, w_i^{T_i}) \in \mathcal{V}_\mathrm{NL}^{T_i}$ denote the corresponding textual description. Both sequences are encoded by the same pretrained LLM $f_{\theta}$: 
$$\mathbf{z}_i^\mathrm{ML}=f_{\theta}(\mathbf{m}_i),  \hspace{0.5cm} \mathbf{z}_i^\mathrm{NL}=f_{\theta}(\mathbf{t}_i)$$ 
The grounding objective encourages the machine token sequence embedding $\mathbf{z}_i^\mathrm{ML}$ to align with its corresponding textual description embedding $\mathbf{z}_i^\mathrm{NL}$, while remaining distinct from embeddings of other items. Given a batch of $N$ items, we employ an InfoNCE-style contrastive loss \cite{InfoNCE}:
\begin{equation}
    \mathcal{L}_i = - \log \frac{\exp\bigl(\mathrm{sim}(\mathbf{z}_i^{\mathrm{ML}}, \mathbf{z}_i^{\mathrm{NL}})/\tau\bigr)}{\sum_{j=1}^N \exp\bigl(\mathrm{sim}(\mathbf{z}_i^{\mathrm{ML}}, \mathbf{z}_j^{\mathrm{NL}})/\tau\bigr)}
    \label{eq:placeholder_label}
\end{equation}
where $sim(\mathbf{u}, \mathbf{v}) = \frac{\mathbf{u}^T\mathbf{v}}{\|\mathbf{u}\|\|\mathbf{v}\|}$ denotes cosine similarity and $\tau$ is a temperature hyperparameter. To promote bidirectional alignment, we also compute the symmetric loss:
\begin{equation}
    \mathcal{L}_i^{sym} = - \log \frac{\exp\bigl(\mathrm{sim}(\mathbf{z}_i^{\mathrm{NL}}, \mathbf{z}_i^{\mathrm{ML}})/\tau\bigr)}{\sum_{j=1}^N \exp\bigl(\mathrm{sim}(\mathbf{z}_i^{\mathrm{NL}}, \mathbf{z}_j^{\mathrm{ML}})/\tau\bigr)}
    \label{eq:placeholder_label}
\end{equation}
The final grounding objective is given by $\mathcal{L} = \frac{1}{2N}\sum_{i=1}^{N}(\mathcal{L}_i + \mathcal{L}_i^{sym})$. 

During grounding, only machine token embeddings are updated; all LLM parameters and natural-language token embeddings remain fixed. Because textual description embeddings are constant, $\mathbf{z}_j^{\mathrm{NL}}$ can be precomputed and reused across training batches, reducing computation. Textual description embeddings are represented using the final hidden state of the sequence, while machine token embeddings are formed via mean pooling over token-level hidden states to ensure equal contribution from each quantization level. 

\subsection{Unified autoregressive modeling}

After grounding machine tokens, we formulate all downstream tasks as sequence-to-sequence generation over heterogeneous token types. Natural-language and machine tokens share a single extended vocabulary $\mathcal{V} = \mathcal{V}_{\mathrm{NL}} \cup \mathcal{V}_{\mathrm{ML}}$ and are modeled autoregressively: for an input sequence $\mathbf{x} = (x_1, \dots, x_T)$ with $x_t \in \mathcal{V}$, the model defines 
\[
p_{\theta}(\mathbf{y}|\mathbf{x}) = \prod_{t=1}^{\|\mathbf{y}\|} p_{\theta}(y_t | \mathbf{x}, y_{<t}), \quad y_{t} \in \mathcal{V} 
\]
No distinction is made between token classes at the modeling level. All tokens share the same embedding table, self-attention layers, and output head, allowing natural-language and machine tokens to interact directly. The framework is task-agnostic. Different downstream tasks are defined by input–output sequence constructions, rather than by changes to the model architecture or training objective.

\begin{figure}[t]
    \centering
    \begin{subfigure}[t]{0.9\linewidth}
        \centering
        \includegraphics[width=\linewidth]{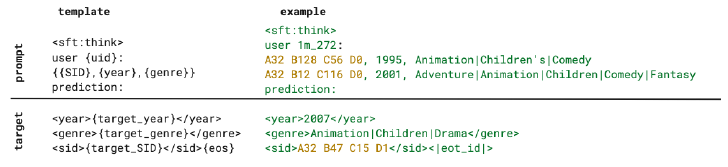}
        \captionsetup{font=scriptsize}
        \caption{Sequential prediction}
        \label{fig:rec_ml_prompt}
    \end{subfigure}


    \begin{subfigure}[t]{1.0\linewidth}
        \centering
        \includegraphics[width=\linewidth]{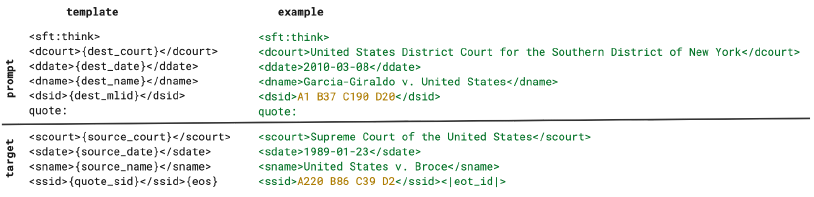} 
        \captionsetup{font=scriptsize}
        \caption{Legal precedent prediction}
        \label{fig:rec_legal_prompt}
    \end{subfigure}

    \captionsetup{font=small}
    \caption{Examples of unified structured prompting across structurally distinct tasks. Each task uses task-specific metadata and a tailored combination of natural-language and machine tokens, yet all follow the same autoregressive SFT training procedure.}
    \label{fig:combined_prompts}
\end{figure}

\subsection{Task formulations}

To operationalize this framework, we represent tasks as structured hybrid sequences that interleave natural-language tokens and machine-native symbols under a consistent template. Inputs follow a fixed formatting schema that organizes heterogeneous fields in a stable order, while outputs are divided into typed segments using delimiter tokens (e.g., \texttt{<year>}, \texttt{<genre>}, \texttt{<sid>}). These output tags define semantic prediction fields and guide autoregressive generation.

\textbf{Sequential recommendation.} We formulate next-item prediction as typed autoregressive generation over heterogeneous fields. A user’s interaction history is encoded as a structured sequence of {\textit{SID}, \textit{year}, \textit{genre}}, where the \textit{SID} is a machine-native identifier and the remaining attributes are natural-language tokens. Rather than directly predicting a single item ID, the model generates the next item in a field-wise manner: it first produces \texttt{<year>} and \texttt{<genre>} segments, followed by \texttt{<sid>}, the machine-native identifier (Figure~\ref{fig:rec_ml_prompt}).

This structured design discourages degenerate solutions that rely solely on machine-token pattern continuation. Instead, it requires the model to align symbolic sequence prediction with natural-language semantics, coupling structural precision with semantic grounding.

\textbf{Legal precedent prediction.} We apply the same typed generative formulation to legal citation modeling. Citation contexts and quoted passages, together with associated metadata, are represented as structured hybrid sequences. Each text segment is encoded and quantized into a machine-native SID, while metadata fields remain in natural-language form. The input prompt interleaves the context SID with its metadata, and the target sequence consists of the cited passage’s metadata followed by its SID (Figure~\ref{fig:rec_legal_prompt}).

This formulation casts citation prediction as structured autoregressive generation over heterogeneous token types. It mirrors the recommendation setting, though it operates over a structurally distinct prediction problem.

%% file: experiment.tex
\subsection{Datasets} 
We evaluate UniLang on six real-world benchmarks covering two structurally distinct domains: sequential recommendation and legal precedent prediction. For sequential recommendation, we use the "Beauty" subset of the Amazon Product Reviews dataset \cite{mcauley2015image}, along with the MovieLens \cite{harper2015movielens} 1M (ML-1m) and 20M (ML-20m) datasets, which differ substantially in size. For legal precedent prediction, we use LePaRD \cite{mahari2024_lepard}, the largest public dataset for legal case retrieval and prediction, which provides three dataset splits of different sizes. Detailed dataset statistics are provided in Appendix \ref{sec:app_dataset}.

\subsection{Evaluation metrics and data splits}
We evaluate model performance using Recall@$k$ and NDCG@$k$. For sequential recommendation, we report results at $k \in {5, 10}$, while for legal precedent prediction we use $k \in {1, 10}$ to emphasize the importance of top-1 accuracy in legal retrieval tasks. For sequential recommendation, we follow the standard leave-one-out protocol, where the last interaction of each user is used for testing, the second-to-last for validation, and the remaining interactions for training. In legal precedent prediction, we use a 90\%/5\%/5\% split for training, validation, and testing, respectively. To enable efficient hyperparameter tuning, we perform model selection on a random subset of the validation set, while all reported test results are computed on the full test set (see Appendix~\ref{sec:val_sampling} for details).

\subsection{Machine-native representation generation details} 
For the Amazon Beauty dataset, we concatenate the raw text from the "title", "categories", "description", "brand", and "price" fields to form a single textual description. For the MovieLens datasets, we combine the "title" with the "genre" field. For the LePaRD dataset, the case context and quoted precedent passages are directly used as provided. Each resulting text blob is encoded using the open-source \texttt{sentence-t5}\footnote{\url{https://huggingface.co/sentence-transformers/sentence-t5-base}}~\cite{ni2022sentence}
 model to obtain a 768-dimensional dense embedding. We then train an RQ-VAE to quantize these embeddings into discrete machine-native codes (SIDs). Detailed training hyperparameters are provided in Appendix~\ref{sec:model_hyper}.

\subsection{Machine token grounding implementation details}
We use \texttt{Llama-3.2-1B-Instruct}~\footnote{\url{https://huggingface.co/meta-llama/Llama-3.2-1B-Instruct}} as the base model and extend its vocabulary with 1,024 machine-native tokens. The embeddings of these tokens are trained using the contrastive objective described in Section~\ref{sec:contrastive_loss} (see Table~\ref{tbl:grounding_hparams} in Appendix for hyperparameters).

Grounding quality is monitored via an alignment metric, computed as the cosine similarity between a machine-token sequence and its corresponding textual embedding. Higher values indicate stronger semantic correspondence. Early stopping is applied based on this metric.

\subsection{LLM fine-tuning details}
After grounding the machine-native tokens and integrating them into the pretrained \texttt{Llama-3.2-1B-Instruct} model, we perform downstream adaptation. During this stage, all original model parameters--including the pretrained token embeddings and the grounded machine-token embeddings--are frozen. We train task-specific Low-Rank Adaptation (LoRA) adapters~\cite{hu2021loralowrankadaptationlarge} using supervised fine-tuning (SFT). For sequential recommendation, the maximum user history length is set to 30 for MovieLens and 50 for Amazon Beauty.

For all tasks, LoRA is applied to the projection layers in each transformer block, namely \texttt{q\_proj}, \texttt{k\_proj}, \texttt{v\_proj}, \texttt{o\_proj}, \texttt{gate\_proj}, \texttt{up\_proj}, and \texttt{down\_proj}. The input–output formatting used for SFT, along with an example prompt and target sequence for MovieLens, is illustrated in Figure~\ref{fig:rec_ml_prompt}. The corresponding formats for LePaRD and Amazon Beauty are shown in Figures~\ref{fig:rec_legal_prompt} and~\ref{fig:beauty_prompt}, respectively. Complete fine-tuning hyperparameters can be found in Table~\ref{tbl:sft_hparams} in the Appendix.

\begin{table}[h]
\centering
\resizebox{\textwidth}{!}{%
\begin{threeparttable}
\captionsetup{font=large, labelfont={large}}
\caption{Performance comparison on sequential recommendation task.}
\label{tab:rec_results}

\begin{tabular}{llccccccr}
\toprule
\multirow{2}{*}{\textbf{Dataset}} 
& \multirow{2}{*}{\textbf{Metric}} 
& \multicolumn{2}{c}{\textbf{Discriminative}} 
& \multicolumn{4}{c}{\textbf{Generative}} 
& \multirow{2}{*}{\textbf{Improvement}} \\
\cmidrule(l){3-5} \cmidrule(l){6-8}
 & 
 &  SASRec
 &  BERT4Rec
 & S\textsuperscript{3}-Rec 
 & P5 
 & \makecell{TIGER \\ (re-impl.)} 
 & \makecell{UniLang \\ (ours)} 
 &  \\
\midrule
\multirow{4}{*}{\textbf{Beauty}} 
 & Recall@5   & \underline{0.0387} & 0.0203  & \underline{0.0387} & 0.0163 & 0.0361 & \textbf{0.0419} & \textcolor{darkgreen}{8.27\%} \\
 & NDCG@5     & \underline{0.0249}  & 0.0124 & 0.0244 & 0.0107 & 0.0241 & \textbf{0.0299} & \textcolor{darkgreen}{20.08\%} \\
 & Recall@10  & \underline{0.0605} & 0.0347  & \textbf{0.0647} & 0.0254 & 0.0595 & 0.0603 & -6.80\% \\
 & NDCG@10    & 0.0318 & 0.0170  & \underline{0.0327} & 0.0136 & 0.0318 & \textbf{0.0358} & \textcolor{darkgreen}{9.48\%} \\
\midrule

\multirow{4}{*}{\textbf{ML-1m}} 
 & Recall@5   & \underline{0.1273} & 0.1011  & 0.1258 & 0.1098 & 0.0634 & \textbf{0.1661} & \textcolor{darkgreen}{30.48\%} \\
 & NDCG@5     & \underline{0.0843}  & 0.0649 & 0.0824 & 0.0734 & 0.0412 & \textbf{0.1145} & \textcolor{darkgreen}{35.82\%} \\
 & Recall@10   & 0.2013 & 0.1533 & \underline{0.2023} & 0.1575 & 0.1055 & \textbf{0.2374} & \textcolor{darkgreen}{17.35\%} \\
 & NDCG@10   & \underline{0.1083}  & 0.0810  & 0.1070 & 0.0888 & 0.0547 & \textbf{0.1376} & \textcolor{darkgreen}{27.05\%} \\
\midrule

\multirow{4}{*}{\textbf{ML-20m}} 
 & Recall@5  & \underline{0.0889}  & 0.0616  & 0.0884 & N/A & 0.0763 & \textbf{0.1911} & \textcolor{darkgreen}{114.96\%} \\
 & NDCG@5    & \underline{0.0549}  & 0.0345  & 0.0531 & N/A & 0.0523 & \textbf{0.1382} & \textcolor{darkgreen}{151.73\%} \\
 & Recall@10  & 0.1453 & 0.0948  & \underline{0.1493} & N/A & 0.1137 & \textbf{0.2597} & \textcolor{darkgreen}{73.95\%} \\
 & NDCG@10   & 0.0728  & 0.0476  & \underline{0.0802} & N/A & 0.0643 & \textbf{0.1603} & \textcolor{darkgreen}{99.88\%} \\

\bottomrule
\end{tabular}

\begin{tablenotes}
\centering
\small
\item Note: The best results are shown in bold, and the second-best results are underlined.
\end{tablenotes}

\end{threeparttable}%
}
\end{table}

\subsection{Performance on sequential recommendation}

We first evaluate UniLang on the sequential recommendation task, comparing it against strong task-specific baselines. BERT4Rec~\cite{sun2019_bert4rec}, SASRec~\cite{SASRec}, and S\textsuperscript{3}-Rec~\cite{zhou2020_s3rec} are discriminative models specifically designed for sequential recommendation. P5~\cite{P5} is a generative approach that verbalizes user, item, and interaction information into natural language, whereas TIGER~\cite{Tiger} is also generative but operates purely on machine-native representations (SIDs), predicting the next SID autoregressively.

For all baselines except TIGER and P5, we report results from the original papers or produce them using the original authors' implementations. For TIGER, we follow the paper and implement the model accordingly. For P5, we adopt corrected and extended results from subsequent works where applicable. Details are provided in Appendix~\ref{sec:sec_rec_baselines}.

As shown in Table~\ref{tab:rec_results}, UniLang achieves remarkable performance, outperforming nearly all baselines across benchmarks with up to 151\% improvement in NDCG@5 and 115\% in Recall@5 on MovieLens-20M. These results demonstrate the effectiveness of extending pretrained LLMs with machine-native symbols: UniLang can directly model and generate structured representations, enabling it to surpass specialized sequential recommendation architectures.

\begin{table}
\centering
\caption{\centering Run-to-run variability on \\MovieLens-20M (mean $\pm$ SE).}
\label{tab:stat_ml20m}
\begin{tabular}{lc}
\toprule
metric & mean $\pm$ standard error \\
\midrule
Recall@5  & 0.1908 $\pm$ 0.00016 \\
NDCG@5    & 0.1378 $\pm$ 0.00017 \\
Recall@10 & 0.2596 $\pm$ 0.00014 \\
NDCG@10   & 0.1600 $\pm$ 0.00014 \\
\bottomrule
\end{tabular}
\end{table}

To assess statistical stability, we perform three independent runs with different random seeds on the MovieLens-20M dataset and report mean ± standard error in Table~\ref{tab:stat_ml20m}.

\begin{table}[h]
\centering
\resizebox{\textwidth}{!}{%
\begin{threeparttable}
\captionsetup{font=large, labelfont={large}}
\caption{ Performance comparison on legal precedent prediction task.}
\label{tab:legal_results}

\begin{tabular}{llcccccr}
\toprule
\multirow{2}{*}{\textbf{Dataset}} 
& \multirow{2}{*}{\textbf{Metric}} 
& \multicolumn{2}{c}{\textbf{Retrieval}} 
& \multicolumn{2}{c}{\textbf{Classification}} 
& \multicolumn{1}{c}{\textbf{Generative}} 
& \multirow{2}{*}{\textbf{Improvement}} \\
\cmidrule(l){3-4} \cmidrule(l){5-6} \cmidrule(l){7-7}
 & & BM25 & \makecell{fine-tuned \\ SBERT} & LEGAL-BERT & DistilBERT & \makecell{UniLang \\ (ours)} & \\
\midrule

\multirow{3}{*}{\textbf{10K}} 
 & Recall@1      & 0.0501 & 0.0899 & 0.1666 & \underline{0.1967} & \textbf{0.2938} & \textcolor{darkgreen}{49.36\%} \\
 & Recall@10     & 0.1952 & 0.4479 & 0.4765 & \underline{0.5912} & \textbf{0.6823} & \textcolor{darkgreen}{15.41\%} \\
 & NDCG@10   & 0.1137 & 0.2627 & 0.3075 & \underline{0.3773} & \textbf{0.4775} & \textcolor{darkgreen}{26.56\%} \\
\midrule

\multirow{3}{*}{\textbf{20K}} 
 & Recall@1      & 0.0413 & 0.0753 & 0.1280 & \underline{0.1674} & \textbf{0.2486} & \textcolor{darkgreen}{48.51\%} \\
 & Recall@10     & 0.1667 & 0.3850 & 0.3684 & \underline{0.5216} & \textbf{0.6290} & \textcolor{darkgreen}{20.59\%} \\
 & NDCG@10   & 0.0956 & 0.2072 & 0.2377 & \underline{0.3291} & \textbf{0.4264} & \textcolor{darkgreen}{29.57\%} \\
\midrule

\multirow{3}{*}{\textbf{50K}} 
 & Recall@1      & 0.0341 & 0.0500 & 0.0877 & \underline{0.1231} & \textbf{0.1676} & \textcolor{darkgreen}{36.15\%} \\
 & Recall@10     & 0.1353 & 0.2590 & 0.2522 & \underline{0.3934} & \textbf{0.4882} & \textcolor{darkgreen}{24.10\%} \\
 & NDCG@10   & 0.0779 & 0.1378 & 0.1625 & \underline{0.2457} & \textbf{0.3131} & \textcolor{darkgreen}{27.43\%} \\

\bottomrule
\end{tabular}

\begin{tablenotes}
\centering
\small
\item Note: The best results are shown in bold, and the second-best results are underlined.
\end{tablenotes}

\end{threeparttable}%
}
\end{table}
\subsection{Performance on legal precedent prediction}
To further demonstrate its generality, we apply the UniLang framework--without any architectural modification--to the structurally different task of legal precedent prediction. Unlike sequential recommendation, which involves modeling chronological sequences of user interactions, legal precedent prediction requires identifying the relevant legal passage from a single citation context, providing a stringent test of UniLang’s ability to generalize across fundamentally different problem structures.

We compare UniLang against established retrieval and discriminative baselines. BM25 and fine-tuned SBERT are retrieval-based methods, whereas LEGAL-BERT and DistilBERT approach passage retrieval as a text classification task. All baseline results are taken from the LePaRD paper \cite{mahari2024_lepard}. Further details on these baselines are provided in Appendix~\ref{sec:legal_baselines}.

Comparison results are provided in Table~\ref{tab:legal_results}. Despite the differences in task structure between sequential recommendation and legal passage retrieval, UniLang consistently outperforms all baselines by a substantial margin. In particular, it achieves a 49\% improvement in Recall@1 on the 10k dataset, significantly surpassing even specialized discriminative models. These results underscore UniLang’s capability to provide a unified modeling framework across tasks with fundamentally different structures.

\begin{figure}[htbp]
    \setlength{\abovecaptionskip}{1pt}  
    \centering
    \begin{minipage}{0.45\textwidth}
        \centering
        \includegraphics[width=\linewidth]{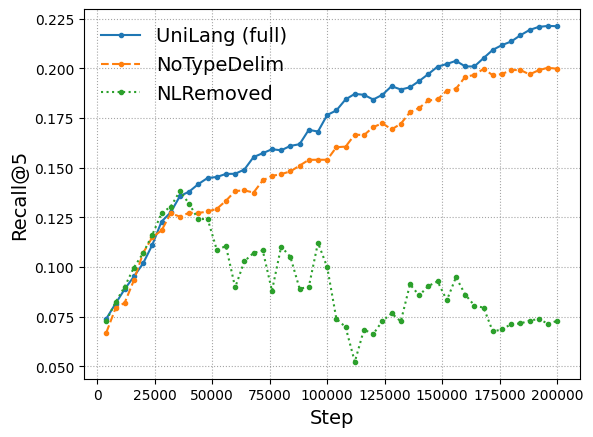}
        \label{fig:fig1}
    \end{minipage}
    \hspace{0.01\textwidth}  
    \begin{minipage}{0.45\textwidth}
        \centering
        \includegraphics[width=\linewidth]{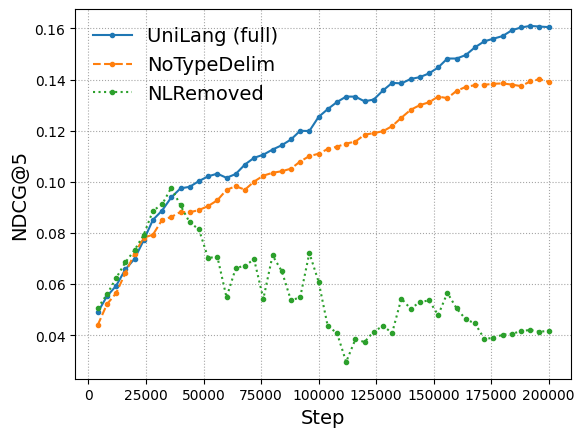}
        \label{fig:fig2}
    \end{minipage}
    \caption{Ablation test on MovieLens-20m. }
    \label{fig:ablation}
\end{figure}

\subsection{Ablation study}
\label{sec:ablation}




We conduct ablation studies on MovieLens-20M to isolate the contributions of semantic grounding, natural-language fields, and structural delimiters. For efficiency, all ablations are evaluated on the fixed validation subset. We consider three variants: (1) NLRemoved, which trains exclusively on machine-native tokens; (2) NoTypeDelim, which removes typed output delimiters (e.g., \texttt{<year>}, \texttt{</dname>}); and (3) NoWarmup, which uses randomly initialized symbolic embeddings instead of those aligned via contrastive learning.

As shown in Figure~\ref{fig:ablation}, NLRemoved exhibits rapid initial improvement followed by performance degradation after 40k steps, suggesting early overfitting when semantic signals from natural language are absent. NoTypeDelim remains stable but consistently underperforms the full UniLang model, indicating that explicit input and output structuring facilitates optimization and improves prediction quality. Notably, NoWarmup fails to achieve non-zero metrics (NDCG and Recall), demonstrating that pre-aligning machine-native symbols to the LLM’s latent space is essential for effective generative learning in our setting. We omit this variant from Figure~\ref{fig:ablation} as it remains at the zero-baseline throughout training. Overall, these results suggest that natural language provides regularizing context, while grounded embeddings and typed structures are critical for stability and prediction quality.


\begin{figure}[htbp]
    \setlength{\abovecaptionskip}{1pt}  
    \centering
    \begin{minipage}{0.95\textwidth}
        \centering
        \includegraphics[width=\linewidth]{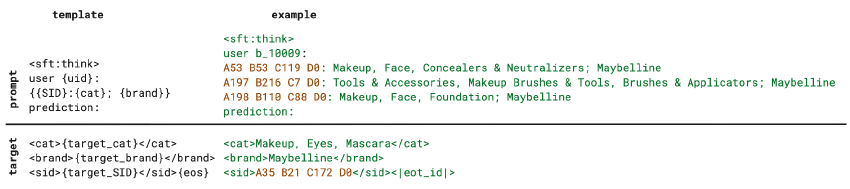}
    \end{minipage}
    \caption{Prompt template and example for the Amazon Beauty dataset. }
    \label{fig:beauty_prompt}
\end{figure}

\subsection{Qualitative comparison}
\label{sec:qualitative_comp}

So far, our results demonstrate the quantitative benefits of jointly representing natural language and machine-native tokens. We now provide a qualitative comparison with approaches that verbalize all user and item information into natural language, such as P5~\cite{xu2024openp5}. P5 typically requires multiple handcrafted prompts per dataset per task (e.g., 13 prompts for Beauty sequential recommendation). One example prompt is shown below:
{\small
\begin{quote}
\textbf{Input template:} I find the purchase history list of user \{\{user\_id\}\}: \{\{history item list of \{\{item\_id\}\}\}\}. I wonder which is the next item to recommend to the user?

\textbf{Target template:} \{\{item [item\_id]\}\}
\end{quote}
}
In contrast, UniLang uses a single structured prompt that seamlessly combines natural language with machine tokens, as shown in Figure~\ref{fig:beauty_prompt}. This unified representation reduces the need for extensive prompt engineering while maintaining or improving performance. By treating language and structured symbols as complementary generative units, UniLang provides a scalable and generalizable framework for heterogeneous tasks, enabling models to jointly model both token types.

\subsection{Emergent representational capabilities}
Beyond task performance, UniLang exhibits an important representational advantage. Existing generative recommender systems struggle with user identity modeling. For example, TIGER~\cite{Tiger} represents users by hashing raw user IDs into a fixed set of 2,000 ID tokens. As a result, these user ID tokens occupy a large portion of the model's extended vocabulary (66\%), which can dominate the token space and limit scalability as the number of users grows.

In contrast, UniLang requires no dedicated user tokens or hashing schemes. User identifiers (e.g., "b\_1009") can be directly represented in natural-language form and processed by the pretrained tokenizer. Although not the primary objective of our method, this property highlights UniLang’s representational scalability and reinforces its role as a unified modeling framework rather than a task-engineered solution.

%% file: conclusion.tex
We introduced \textit{UniLang}, a unified framework that extends pretrained LLMs to model natural language and machine-native symbols within a single autoregressive objective. By treating structured symbols as first-class generative units, UniLang enables structurally diverse tasks--such as sequential recommendation and legal precedent prediction--to be addressed without task-specific architectures. Empirically, UniLang consistently outperforms strong baselines, while ablations demonstrate the importance of natural-language context and typed structure. Overall, UniLang provides a general interface for extending pretrained LLMs beyond language to model heterogeneous machine-native representations.

While UniLang demonstrates strong performance on structured prediction tasks, we did not evaluate the impact of symbolic fine-tuning on the model’s natural language generation quality. Our focus in this work is on accuracy in predictive tasks, and studying potential effects on language fluency and coherence is left for future work.

%% file: appendix.tex
\section{Dataset}
\label{sec:app_dataset}

\begin{table}[t]
\centering
\small
\setlength{\tabcolsep}{0pt} 
\renewcommand{\arraystretch}{1.2}
\caption{Statistics of the sequential recommendation datasets.}
\label{tbl:rec_stat}

\begin{tabular*}{\textwidth}{@{\extracolsep{\fill}}lrrrrr}
\toprule
Dataset & \#users & \#items & \#actions & Avg. length & Density \\ 
\midrule
Beauty         & 40,226  & 54,542  & 0.35m   & 8.8   & 0.02\% \\
ML-1m   & 6,040   & 3,416   & 1m & 163.5 & 4.79\% \\
ML-20m  & 138,493 & 26,744  & 20m& 144.4 & 0.54\% \\
\bottomrule
\end{tabular*}
\end{table}

\begin{table}[t]
\centering
\small
\renewcommand{\arraystretch}{1.2}
\caption{Statistics of the legal precedent prediction datasets.}
\label{tbl:lepard_stat}

\begin{tabular}{rrr}
\toprule
\makecell{Dataset \\ (\# cited passages) }
& \makecell{\# Citing \\ Passage Surface Forms} 
& \makecell{\# Cited \\ Passage Surface Forms} \\
\midrule
10k  & 1,294,730 & 798,558 \\
20k  & 1,586,663    & 1,140,058 \\
50k  & 2,079,704    & 1,821,751 \\
\bottomrule
\end{tabular}
\end{table}

\begin{table}[t]
\centering
\small
\renewcommand{\arraystretch}{1.2}
\caption{Summary statistics of legal precedent dataset text features}
\label{tbl:lepard_text_stat}

\begin{tabular}{lrrrr}
\toprule
Feature & Mean & Std & Min & Max  \\
\midrule
Length of cited text (chars)  & 306 & 225 & 24 & 18,342\\
Length of citing context (chars)    & 562 & 216 & 5 & 14,062 \\
\bottomrule
\end{tabular}
\end{table}

\begin{itemize}
    \item Amazon Beauty~\footnote{\url{https://cseweb.ucsd.edu/~jmcauley/datasets/amazon/links.html}}~\cite{mcauley2015image}: This dataset consists of product reviews collected from Amazon.com. The full collection is organized by top-level product categories, and in our experiments, we use the subset corresponding to the "Beauty" category.
    \item MovieLens: A widely used benchmark for evaluating recommendation systems. In this study, we use two standard versions: MovieLens 1M (ML-1m)~\footnote{\url{https://grouplens.org/datasets/movielens/1m/}} and MovieLens 20M (ML-20m)~\footnote{\url{https://grouplens.org/datasets/movielens/20m/}}.
    \item The LePaRD dataset~\footnote{\url{https://huggingface.co/datasets/rmahari/LePaRD}}~\cite{mahari2024_lepard} is a large-scale collection of 4.3 million U.S. federal judicial citations. It leverages judges' quotation behavior as supervision, pairing precedential quotations with their corresponding citation contexts. The dataset provides three subsets that differ in the number of top-$n$ most frequently cited passages included. Due to unavoidable preprocessing noise—such as OCR errors, sentence segmentation inconsistencies, and metadata formatting variations—the same citing context or cited quotation may appear in multiple surface forms. 
\end{itemize}
Table~\ref{tbl:rec_stat} presents the statistics of the sequential recommendation datasets. Table~\ref{tbl:lepard_stat} provides an overview of the legal precedent prediction dataset. Table~\ref{tbl:lepard_text_stat} summarizes the statistics of the textual features in the legal dataset.

\section{Validation set sampling}
\label{sec:val_sampling}
\begin{table}[t]
\centering
\small
\renewcommand{\arraystretch}{1.2}
\caption{Validation subset sizes for model selection}
\label{tbl:validation_sample}

\begin{tabular}{crr}
\toprule
Dataset & Validation size & Sample size  \\
\midrule
Beauty & 40,226  & 5,000\\
MovieLens-1m & 6,040 & 1,000\\
MovieLens-20m & 138,493 & 1,000\\
10k & 103,812 & 1,000 \\
20k & 134,737 & 1,000 \\
50k & 190,051 & 1,000 \\
\bottomrule
\end{tabular}
\end{table}

We use Recall@$k$ on the validation set for early stopping during training. Since our framework is generative, we perform beam search to deterministically generate the top-$k$ results and compute Recall@$k$ accordingly. Running beam search over the full validation set is computationally expensive, so to expedite model selection, we instead use a fixed, randomly sampled subset of the validation set generated with a fixed random seed. For all datasets, reported results are based on models selected using this sampled subset. Table~\ref{tbl:validation_sample} lists the sample sizes for each dataset. For all experiments, we use a beam size of 20.

\section{Model training hyperparameters}
\label{sec:model_hyper}

All experiments were conducted on 8 NVIDIA H100 GPUs (80GB each). Training can also be performed on a single GPU; multiple GPUs are used primarily to accelerate training via data parallelism. Training time varies depending on batch composition. On the MovieLens-20M dataset, under our configuration, training proceeds at approximately 0.15 seconds per optimization step (batch size 128), with total runtime scaling linearly with the number of training steps.

\begin{table}[t]
\centering
\small
\setlength{\tabcolsep}{3pt}      
\renewcommand{\arraystretch}{1.1}
\captionsetup{font=small}
\caption{RQ-VAE hyperparameters.}
\label{tbl:ravae_hparams}

\begin{tabular*}{\textwidth}{@{\extracolsep{\fill}}lccc}
\toprule
Hyperparameter
& Beauty
& \makecell{ML-1M \\ ML-20M} 
& \makecell{LePaRD \\ 10k / 20k / 50k} \\
\midrule
Total training steps &30k & 20k & 20k \\
Warm up  &3k & 2k & 3k \\
Peak learning rate   &1e-3 & 1e-3 & 1e-3  \\
End learning rate  &1e-5 & 5e-5  &  1e-5 \\
Ema decay & 0.99    & 0.99   & 0.99 \\
Commitment cost & 1.5 & 1.5 & 0.1 \\
\bottomrule
\end{tabular*}

\vspace{3pt}
\footnotesize
\parbox{\textwidth}{\raggedright
\textit{Note:} All experiments use a codebook embedding size of 16 and a batch size of 2048, and are optimized with AdamW (weight decay $0.055$, $\beta_1 = 0.9$, $\beta_2 = 0.98$) under a cosine learning rate schedule.
}
\end{table}

\begin{figure}[htbp]
    \setlength{\abovecaptionskip}{1pt}
    \centering

    \begin{subfigure}[t]{0.45\textwidth}
        \centering
        \includegraphics[width=\linewidth]{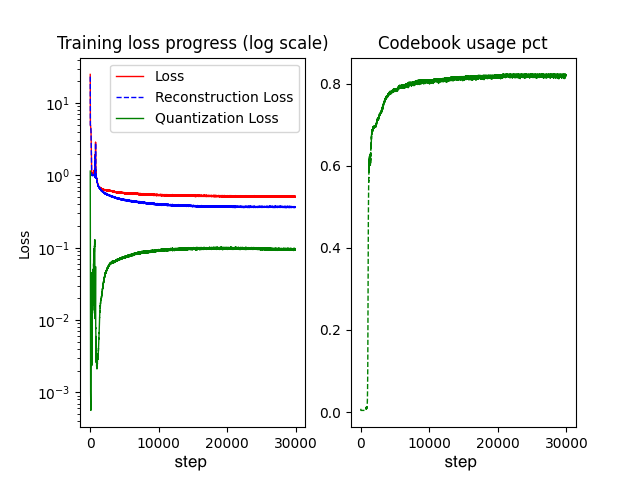}
        \caption{Beauty}
        \label{fig:rqvae_beauty}
    \end{subfigure}
    \hspace{0.02\textwidth}
    \begin{subfigure}[t]{0.45\textwidth}
        \centering
        \includegraphics[width=\linewidth]{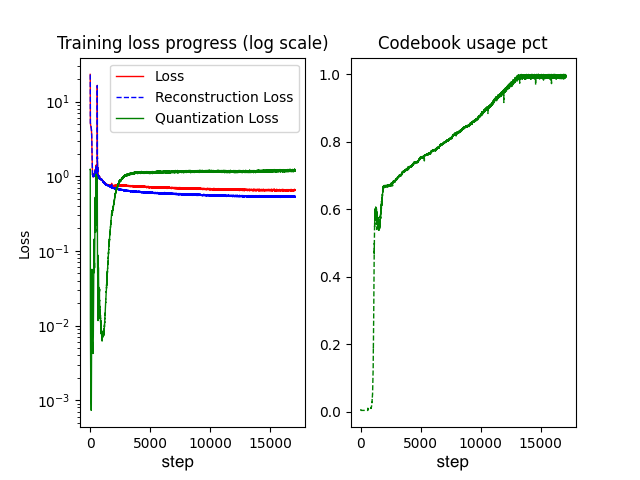}
        \caption{LePaRD 20k}
        \label{fig:rqvae_lepard}
    \end{subfigure}

    \vspace{4pt}   
    \caption{RQ-VAE training progress on different datasets.}
    \label{fig:rqvae_training}
\end{figure}

\subsection{RQ-VAE training parameters} The same RQ-VAE architecture is used across all datasets. The model takes 768-dimensional text embeddings as input. The encoder comprises three fully connected layers with output dimensions 512, 256, and 128, each followed by ReLU activation, and outputs a 16-dimensional latent representation. The decoder mirrors the encoder with layers of dimensions 128, 256, and 512. At each quantization level, we maintain a codebook of size 256. Table~\ref{tbl:ravae_hparams} summarizes the detailed hyperparameter settings, and Figure~\ref{fig:rqvae_training} illustrates the training dynamics of the individual loss components across datasets.

\begin{table}[t]
\centering
\small
\setlength{\tabcolsep}{3pt}
\renewcommand{\arraystretch}{1.1}
\captionsetup{font=small}

\caption{Machine token alignment hyperparameters.}
\label{tbl:grounding_hparams}

\begin{tabular*}{\textwidth}{@{\extracolsep{\fill}}lccc}
\toprule
Dataset
& Beauty
& \makecell{ML-1M \\ ML-20M} 
& \makecell{LePaRD \\ 10k / 20k / 50k} \\
\midrule
Batch size    & 1024 & 512 & 128 \\
Learning rate & $1\mathrm{e}{-3}$ & $1\mathrm{e}{-3}$ & $6\mathrm{e}{-3}$ \\
Total steps   & 4k   & 4k  & 8k \\
\bottomrule
\end{tabular*}

\vspace{3pt}
\footnotesize
\parbox{\textwidth}{\raggedright
\textit{Note:} All experiments use the AdamW optimizer (weight decay $10^{-2}$, $\beta_1 = 0.9$, $\beta_2 = 0.98$) with a cosine learning rate schedule and 400 warmup steps. The temperature for the contrastive loss is set to 0.2.

}
\end{table}

\subsection{Machine token grounding parameters}
Table~\ref{tbl:grounding_hparams} summarizes the hyperparameters used for training the embeddings of the machine-native tokens.

\begin{table}[t]
\centering
\small
\setlength{\tabcolsep}{3pt}      
\renewcommand{\arraystretch}{1.1}
\setlength{\abovecaptionskip}{0pt}
\setlength{\belowcaptionskip}{2pt}
\captionsetup{font=small}
\caption{SFT hyperparameters.}
\label{tbl:sft_hparams}

\begin{tabular*}{\textwidth}{@{\extracolsep{\fill}}lccc}
\toprule
Dataset
& Beauty
& \makecell{ML-1M \\ ML-20M}
& \makecell{LePaRD \\ 10k / 20k / 50k} \\
\midrule
Batch size    & 8 & 32/128 & 512 \\
Learning rate & $1\mathrm{e}{-4}$ & $2\mathrm{e}{-4}$ & $2\mathrm{e}{-4}$ \\
LoRA dropout  & 0.25  & 0.05  & 0.25 \\
Total training steps  & 30k  & 20k & 30k \\
Warm up & 2k & 2k & 2k \\
\bottomrule
\end{tabular*}

\vspace{3pt}
\footnotesize
\parbox{\textwidth}{\raggedright
\textit{Note:} All experiments use LoRA with rank 32 and $\alpha=2$. Optimization is performed using AdamW (weight decay $0.005$, $\beta_1 = 0.9$, $\beta_2 = 0.98$) with a cosine learning rate schedule.
}
\end{table}

\subsection{Supervised fine tuning parameters}
Table~\ref{tbl:sft_hparams} lists the hyperparameters used for supervised fine-tuning of \texttt{Llama-3.2-1B-Instruct} on structured input–output sequences containing interleaved natural-language and machine-native tokens.

\section{Sequential recommendation baseline methods}
\label{sec:sec_rec_baselines}
\begin{itemize}
    \item SASRec~\cite{SASRec} is a self-attention-based sequential recommendation model that efficiently captures long-term user behavior. 
    
    \item BERT4Rec~\cite{sun2019_bert4rec} trains a bidirectional Transformer with a masked item prediction objective to learn contextualized representations of user behavior sequences.

    \item S\textsuperscript{3}-Rec~\cite{zhou2020_s3rec} adopts a self-supervised pretraining strategy to capture correlations among items, attributes, and interaction sequences, alleviating data sparsity.

    \item P5~\cite{P5} formulates recommendation as a language modeling problem, representing user–item interactions, user profiles, item metadata, and reviews in natural language.

    \item TIGER~\cite{Tiger} represents items using semantic IDs (SIDs), converting user interaction histories into SID sequences. A generative model is trained from scratch to autoregressively predict the next SID for sequential recommendation.
\end{itemize}

In evaluating recommendation systems, two protocols are commonly adopted: (1) sampled evaluation, where the ground-truth item is ranked against a fixed set of negative samples (e.g., 99 items), and (2) full-ranking, where the ground-truth item is ranked against all items in the candidate set. While sampled evaluation is computationally efficient, it can lead to overly optimistic performance estimates. In this work, we adopt full-ranking evaluation throughout to provide a more rigorous and realistic assessment of model performance.

For the Amazon Beauty dataset, results for SASRec, BERT4Rec, and S\textsuperscript{3}-Rec are taken from the publicly released benchmarks~\footnote{\url{https://github.com/RUCAIBox/CIKM2020-S3Rec}} provided by the S\textsuperscript{3}-Rec authors. The P5 results are obtained from the TIGER paper~\cite{Tiger}, where the authors introduced a preprocessing modification to ensure a fair comparison.

For the MovieLens 1M and 20M datasets, results for SASRec, BERT4Rec, and S\textsuperscript{3}-Rec are obtained using the authors’ released code. All models are trained with their default hyperparameter settings. For SASRec and BERT4Rec, we additionally implement full-ranking evaluation to obtain the reported results.

Results for P5 on the MovieLens 1M dataset are taken from the OpenP5~\cite{xu2024openp5} paper. OpenP5 incorporates  the original P5 backbone model, downstream tasks, and item indexing method. Since P5 requires substantial manual prompt engineering when adapted to new datasets, we consider using OpenP5's reported results to provide a fairer comparison.

Due to the prohibitively high storage and computational cost of applying P5 to the MovieLens 20M dataset, we do not report its results on this dataset (denoted as N/A in Table~\ref{tab:rec_results}).

\section{Legal precedent prediction baseline methods}
\label{sec:legal_baselines}

We consider four baseline methods, two retrieval-based and two classification-based.  

BM25 represents a sparse lexical retrieval approach~\cite{robertson1994okapi}. The "fine-tuned SBERT" baseline is a dense embedding-based retrieval method that uses a fine-tuned SBERT~\cite{reimers2019sentencebert} model to generate embeddings, followed by maximum dot-product similarity for retrieval.  

Passage retrieval can also be formulated as a text classification task, where each target passage is assigned a unique label that serves as the prediction target for its preceding context. Two classification-based baselines are considered: DistilBERT~\cite{sanh2020distilbert} and LEGAL-BERT~\cite{chalkidis2020legalbert}, the latter being a domain-adapted BERT model trained on a large corpus of legal documents. All baseline results are taken from the LePaRD paper~\cite{mahari2024_lepard}.